\documentclass[final]{agujournal}

\usepackage{amsmath, amssymb}

\usepackage{todonotes}
\usepackage{tikz}
\usetikzlibrary{calc}
\definecolor{myred}{rgb}{0.761, 0.0, 0.1}
\definecolor{myblue}{rgb}{0.0, 0.376, 0.635}
\definecolor{mygreen}{rgb}{0.0, 0.620, 0.176}
\definecolor{mypurple}{rgb}{0.553, 0.133, 0.537}

\usepackage{pgfplots}
\usepgfplotslibrary{groupplots}

\usepackage{contour}

\newcommand{\change}[1]{{#1}}

\newcommand{\morechange}[1]{{#1}}

\journalname{Geophysical Research Letters}

\usepackage{siunitx}
\DeclareSIUnit\hour{hr}

\usepackage{etoolbox}
\newtoggle{gentikz} 
\newtoggle{usetikz}

\togglefalse{gentikz}
\togglefalse{usetikz}

\iftoggle{gentikz}{%
	\usetikzlibrary{external}
	\tikzexternalize 
	\toggletrue{usetikz}
}{}

\begin{document}

%
%


\title{Remote measurement of sea ice dynamics with regularized optimal transport}

%
%




\authors{M. D. Parno\affil{1}\thanks{72 Lyme Road, Hanover, NH 03741},
  B. A. West\affil{1},
  A. J. Song\affil{1},
  T. S. Hodgdon\affil{1},
  and D. T. O'Connor\affil{1}}


\affiliation{1}{U.S. Army Corps of Engineers, Cold Regions Research and Engineering Laboratory, Hanover, NH}




\correspondingauthor{Matthew D. Parno}{Matthew.D.Parno@usace.army.mil}




\begin{keypoints}
\item A novel approach is proposed that can measure sea ice motion in the presence of topological changes like break up and fracture.
\item Techniques from optimal transport are used to estimate sea ice velocities and strain at the native resolution of an image pair.
\item Results using MODIS imagery of Nares Strait illustrate the ability of our approach to provide high resolution estimates of ice motion that agree with expected ice behavior.
\end{keypoints}

%
%


\begin{abstract}
As Arctic conditions rapidly change, human activity in the Arctic will continue to increase and so will the need for high-resolution observations of sea ice.  While satellite imagery can provide high spatial resolution, it is temporally sparse and significant ice deformation can occur between observations.   This makes it difficult to apply feature tracking or image correlation techniques that require persistent features to exist between images. \change{With this in mind, we propose a technique based on optimal transport, which is commonly used to measure differences between probability distributions.  \morechange{When little ice enters or leaves the image scene, we show that} regularized optimal transport can be used to quantitatively estimate ice deformation.  We discuss the motivation for our \morechange{approach} and describe efficient computational implementations.  Results are provided on a combination of synthetic and MODIS imagery to demonstrate the ability of our approach to estimate dynamics properties at the original image resolution.}
\end{abstract}

%
%

%


%
%
%
%

\section{Introduction}\label{sec:intro}
Sea ice plays a significant role in the Earth's climate system and directly impacts human activities in the Arctic. Any characteristic change in the ice will therefore have major global impacts; climatologically, economically, and militarily.  Recent observations have shown a rapid decline in seasonal ice thickness and extent, as well as a wider band of unconsolidated ice along the pack edge~\citep{screen2010central, jeffries2013arctic, carmack2015toward, strong2013arctic}. We do not, however, have a full understanding of why these significant changes are occurring. Models of sea ice, which represent our quantitative understanding of ice evolution, have continuously underestimated the rapid decline of Arctic sea ice~\citep{stroeve2012trends} and generally disagree on the spatial distribution of ice~\citep{dukhovskoy2015skill}.  This suggests that the models are missing key information about how the ice pack evolves.

%
%
%

We believe that the discrepancy between models and observations is primarily caused by an inadequate understanding and subsequent mathematical characterization of the ice dynamics. The ice evolution is controlled by both the ice dynamics and thermodynamics, but we believe the relatively fast changing dynamics are not understood as well as the slower thermodynamics and thus contribute more to the mismatch between model predictions and observations. High fidelity models, such as those based on the discrete element method, have the potential to more accurately capture dynamic events like cracking and ridging~\citep{herman2016discrete, hopkins1998four, hopkins2004formation}, but require high resolution imagery for validation, the construction of realistic initial conditions, and the calibration of contact laws.  To improve our understanding of the ice dynamics and validate high fidelity models, we therefore need high resolution observations of the dynamics, including position, velocity, and other variables such as strain.

Previous approaches for computing ice velocities, including cross correlation methods \citep{ninnis1986automated, lavergne2010sea, komarov2014sea, lindsay2003} such as normalized cross correlation (NCC), have typically relied on forms of windowing that homogenize the original high resolution image and subsequently limit the resolution of derived velocity fields. Despite their shortcomings, these methods are commonly employed with synthetic aperture radar (SAR) and visible electromagnetic optical (EO) imagery to provide ice velocity estimates.
%
%
NCC benefits from relatively easy implementation and computational efficiency, but relies on persistent features in image pairs and thus suffers when displacements are large, the material shears significantly, or when breakup occurs and the ice topology does not persist from image to image. Therefore, the cross-correlation must be computed over a \change{region of interest (ROI)} that is large enough to contain these persistent features. This effectively downsamples the NCC derived velocity field from the original image resolution and therefore loses fine-scale features of the image. \change{There has also} been recent work using optical flow to estimate ice velocities with high spatial resolution \change{\citep{petrou2017high,petrou2018towards}}. This is a promising approach, but struggles with temporal variations in pixel intensity, which is common with satellite observations, \change{and requires smoothness assumptions on the velocity field, which may not be valid when cracks are present}. Both NCC and optical flow techniques can be difficult to apply on real-world images, especially those in the marginal ice zone, where it is common to see temporal intensity variations, breakup, shear, and rotation. 

With these challenges in mind, we propose a new technique for analyzing temporal changes in sea ice imagery and extracting dynamic properties like velocity and strain.  Our approach is based on the Wasserstein metric from optimal transport theory.  When applied to satellite imagery, this metric provides both a global measure of image similarity as well as a mechanism for identifying the most dynamically active regions of the image, estimating the ice velocity field, and extracting the strain field. These measurements are made without the need for feature detection or localized cross correlation, thus reducing information lost by windowing and making them robust in break up scenarios.  \morechange{The accuracy of our approach is tied to the amount of ice that enters or leaves the image scene, but can otherwise be applied with minimal imagery preprocessing.   In Sections \ref{sec:wass} and \ref{sec:conclusion} we describe this conservation assumption in more detail.}

Section \ref{sec:wass} describes the Wasserstein metric in greater detail and provides background for additional concepts used in our method. Then in Section \ref{sec:sea_ice} we describe how these concepts are used to analyze changes in sea ice imagery. We illustrate the effectiveness of this method on synthetic images and MODIS imagery of the Nares Strait in Sections \ref{sec:results:synthetic} and \ref{sec:results:nares}, respectively. Lastly, we provide concluding remarks in Section \ref{sec:conclusion}.

\section{Wasserstein distance background}\label{sec:wass}
In order to use the mathematical machinery of optimal transport, we will restrict our attention to single band images and treat them as normalized probability mass functions.  The difference between images can then be measured using a statistical metric called the Wasserstein distance.  Computing the Wasserstein distance between two images involves constructing a transformation between the images.  \change{Interestingly}, dynamic variables like velocity and strain can be extracted directly from this transformation.

To see this mathematically, consider two \change{one dimensional} vectors $\tilde{p}$ and $\tilde{q}$ with $N=N_x N_y$ components that represent the intensities of all pixels in two different images of size $N_x \times N_y$. We will refer to $\tilde{p}$ as the source image and $\tilde{q}$ as the target image, which in the applications below will correspond to remote sensing observations over the same region on two different days.   \change{Note that each pixel in an image corresponds to a component in the vector and we will therefore refer to components and pixels interchangeably.} Normalizing these vectors to ensure they sum to $1$, we obtain two valid probability mass functions $p$ and $q$ defined by
\begin{linenomath*}
\begin{equation}
p = \frac{\tilde{p}}{\sum_{i=1}^N \tilde{p}_i}, \quad
q = \frac{\tilde{q}}{\sum_{i=j}^N \tilde{q}_j}.
\end{equation}
\end{linenomath*}
Our goal is to use this probabilistic interpretation to measure the difference between the normalized images $p$ and $q$ with techniques from statistics.  

There are many metrics for comparing probability distributions, including the total variation distance, Kullback-Leibler divergence, generalized f-divergences, and the Wasserstein distance.  Our focus will be on approximating the Wasserstein distance, which is physically meaningful as the solution of a potential flow equation \citep{Benamou2000}, has been rigorously studied theoretically \citep{Brenier1991,Villani2008}, and can be efficiently computed with recent algorithmic advancements \citep{Cuturi2013, solomon2015, Benamou2015}. \change{The connection between optimal transport and fluid flow has been explored for particle image velocimetry \citep{agueh2015optimal, saumier2015optimal}, but to our knowledge has not previously been applied to satellite imagery.   By implicitly assuming irrotational flow \citep{Benamou2000}, our optimal transport approach is not expected to accurately characterize pure rotational deformation.  However, the results of \cite{saumier2015optimal} and our results below indicate that optimal transport can work well in practice despite this limitation.}


To define the Wasserstein distance, consider the set $\Pi(p,q)$ of positive $N\times N$ matrices whose row and column sums are $p$ and $q$, respectively.  This set describes the possible joint distributions or ``couplings", denoted by $\gamma$, between the source image $p$ and target image $q$.
Thus, for any coupling $\gamma \in \Pi(p,q)$, the row and column sums satisfy $\sum_j \gamma_{ij} = p_i$ and $\sum_i \gamma_{ij} = q_j$.   Notice that row $i$ of a coupling $\gamma$ also defines a mapping from pixel $i$ in the source image $p$ to one or more pixels in target image $q$.  A natural way to measure the ``cost" associated with the coupling $\gamma$ is thus to measure the average distance that mass in the source image $p$ is moved when it is transformed into the target image $q$.  To make this concept more concrete, consider a ``ground cost" $c_{ij}$ that characterizes the ``cost" or ``work" required to move one unit of mass from pixel $i$ to pixel $j$.  In practice, $c_{ij}$ will correspond to the squared Euclidean distance between the center of pixel $i$ and the center of pixel $j$.
The overall transport cost of a coupling $\gamma$ is then given by $\sum_{ij} \gamma_{ij} c_{ij}$, i.e., the sum of the transport cost between pixels $i$ and $j$ multiplied by the probability $\gamma_{ij}$ of moving from pixel $i$ to pixel $j$. 

 \textit{The Wasserstein distance between $p$ and $q$, denoted by $W(p,q)$, is the minimum transport cost obtained by any coupling in $\Pi(p,q)$.}  More precisely, the Wasserstein distance is given by the linear optimization problem
\begin{linenomath*}
\begin{equation}
W(p,q) = \underset{\gamma \in \Pi(p,q)}{\text{min}}  \sum_{i,j=1}^N c_{ij} \gamma_{ij}. \label{eq:standardWass}
\end{equation}
\end{linenomath*}
Unfortunately, even state of the art algorithms for solving \eqref{eq:standardWass} directly have $O(N^3)$ complexity \citep{Pele2009,Cuturi2013} and are thus computationally intractable for high resolution images containing a large number of pixels $N$.  However, as shown by \cite{Cuturi2013} and subsequent works, adding a regularization term based on the entropy of the coupling $\gamma$ leads to a faster algorithm with linear convergence rates.


The entropy-regularized Wasserstein \change{distance} will be denoted by $W_{\epsilon}(p,q)$ and is defined by
\begin{linenomath*}
\begin{equation}
W_\epsilon(p,q) = \underset{\gamma \in \Pi(p,q)}{\text{min}}  \sum_{i,j=1}^N c_{ij} \gamma_{ij} - \epsilon H(\gamma), \label{eq:RegularizedOT}
\end{equation}
\end{linenomath*}
where $\epsilon$ is a relaxation parameter and $H(\gamma) = -\sum \gamma_{ij}\log(\gamma_{ij})$ is the statistical entropy of $\gamma$. The addition of the entropy term $- \epsilon H(\gamma)$ makes the optimization problem strongly convex and thus easier to solve. 
 Larger relaxation parameters $\epsilon$ result in more diffuse optimal couplings with larger entropies, which means the regularized Wasserstein distance will be dominated by the entropy term and will lose its utility as a metric between $p$ and $q$.  Fortunately, only a small value of $\epsilon$ is needed in practice when using the algorithms of \cite{Cuturi2013} and \cite{Benamou2015}.  \change{As described below,} using these algorithms it is possible to efficiently solve \eqref{eq:RegularizedOT} even with high resolution remote sensing images.  

\cite{Cuturi2013} noticed that under mild technical conditions, the coupling $\gamma^{(n)}$ can be described by rescaling the rows and columns of a matrix $\xi = \exp\left[ - c/ \epsilon \right]$ to obtain
\begin{linenomath*}
\begin{equation}
\gamma^{(n)} = \text{diag}\left(u^{(n)}\right) \xi  \text{diag}\left( w^{(n)} \right), \label{eq:diagonalScaling}
\end{equation}
\end{linenomath*}
where $\text{diag}(x)$ denotes a matrix with the vector $x$ along the diagonal, and the vectors $u^{(n)}, w^{(n)}\in\mathbb{R}^N$ satisfy the recursive relationship
\begin{linenomath*}
\begin{equation}
u^{(n)} = \frac{p}{\xi w^{(n)}} \quad \text{ and } \quad w^{(n+1)} = \frac{q}{\xi^T u^{(n)}}, \label{eq:sinkhornIterates}
\end{equation}
\end{linenomath*}
where division is taken componentwise.  As $n\rightarrow \infty$, the coupling $\gamma^{(n)}$ convergences to the optimal coupling $\gamma^\ast$ with linear convergence rates obtained for appropriate costs and distributions \citep{Cuturi2013, Benamou2015}.   Thus, to compute $\gamma^\ast$, we start with an arbitrary $u^{(0)}$ and iterate until the change in $u$ and $w$ is small, e.g., $\| u^{(n)} - u^{(n-1)} \|<10^{-6}$, or a maximum number of iterations is reached, e.g., $n=10^3$.  The converged values of $u$ and $v$ will be denoted as $u^\ast$ and $v^\ast$.   We choose $u^{(0)}$ to be a vector of all ones.

The ground cost $c_{ij}$ plays a critical role in the Wasserstein distance $W_\epsilon(p,q)$.  A common choice for the ground cost is the squared Euclidean distance $c_{ij}=\|x_i-x_j\|^2$, which has deep ties to movement of mass in a potential flow field \citep{Benamou2000}.  \change{However, using the ground costs to build the kernel matrix $\xi$ directly can become intractable for large remote sensing images with potentially millions of pixels.  To overcome this, we have adopted the convolutional approach of \cite{solomon2015}, which uses the relationship between distance and the heat equation defined by Varadhan's theorem \citep{Crane2013} to compute the action of the kernel matrix $\xi$ on a vector using only Gaussian convolutions.  These convolutions can be quickly computed with limited memory usage and are available in many existing image processing toolboxes.}

\section{Application to sea ice imagery}\label{sec:sea_ice}
The Wasserstein distance $W_\epsilon(p,q)$ in \eqref{eq:RegularizedOT} is a global measure of the difference between $p$ and $q$.  This is valuable information in itself, but the optimal coupling $\gamma^\ast$ solving \eqref{eq:RegularizedOT} can also be used to highlight regions of significant change, approximate the ice velocity, and estimate strain in the sea ice.  Recall that $c_{ij}$ represents the cost of moving mass from pixel $i$ to pixel $j$ and $\gamma_{ij}^\ast$ represents the amount of mass that is moved from pixel $i$ to pixel $j$.  The average cost of transporting the mass from pixel $i$ is then given by the partial sum
\begin{linenomath*}
\begin{eqnarray}
\bar{c}_i &=& \frac{1}{p_i} \sum_{j} \gamma_{ij}^\ast c_{ij} \label{eq:magnitudeApprox},
\end{eqnarray}
\end{linenomath*}
which we call the transport distance for pixel $i$.  Note that the expression in \eqref{eq:magnitudeApprox} is a conditional expectation over the cost associated with pixel $i$ and that the $1/p_i = 1/(\sum_{j}\gamma_{ij}^\ast)$ term is needed to normalize the $i^{th}$ row of $\gamma^{\ast}$.  The transport distance is an indication of how much the sea ice in pixel $i$ is moving between images $p$ and $q$.   When the source image $p$ and target image $q$ are different snapshots of the same region, the transport cost becomes a measure of ice deformation and can be used to study both short term ice dynamics as well longer seasonal trends or deformation climatologies.

Recall that the coupling $\gamma^\ast$ provides a stochastic description of the transformation between the source image $p$ and the target images $q$.  For calculating velocities and strain rates however, a one-to-one deterministic mapping from pixels in $p$ to pixels in $q$ is desired.  One such deterministic map is the barycentric projection map defined in \cite{peyre2017computational}.  Let $(x^p_i, y^p_i)$ denote the 2D location of pixel $i$ in the source image $p$ and let $(x^q_i, y^q_i)$ be the location where the mass from $p_i$ is transported to in the target image $q$.  The barycentric projection map defines $x^q$ and $y^q$ as 
\begin{linenomath*}
\begin{eqnarray}
x^q = \frac{\xi (w^\ast  \odot x^p ) \odot u^\ast}{p} \label{eq:transportPlan1} \\ 
y^q = \frac{\xi (w^\ast  \odot y^p ) \odot u^\ast}{p}, \label{eq:transportPlan2} 
\end{eqnarray}
\end{linenomath*}
where $\odot$ represents the componentwise product. The transformation $(x^p_i, y^p_i) \rightarrow  (x^q_i, y^q_i)$ is thus known for every pixel $i$.   If $p$ is an image obtained at time $t$ and $q$ is an image at time $t+\delta t$, then 
\begin{linenomath*}
\begin{equation}
v_i = \left[ \begin{array}{c} \frac{x_i^q-x_i^p}{\delta t}\\  \frac{y_i^q - y_i^p}{\delta t} \end{array}\right],  \label{eq:velocityApprox}
\end{equation}
\end{linenomath*}
 is an estimate of the velocity for pixel $i$ in the image.

The spatial derivatives of the velocity make up the strain rate tensor and multiplying by $\delta t$ gives an incremental strain $\varepsilon$.  Thus, finite difference derivatives (in space) of the velocity vector $v_i$ can be used to estimate the partial derivatives in the Jacobian matrix $\nabla v_i$.  We use a second order central finite difference scheme.  An approximation of the incremental strain tensor at pixel $i$ is then given by \change{a finite difference approximation of the symmetric gradient}
\begin{linenomath*}
\begin{equation}
\varepsilon_i = \frac{\delta t}{2}\left( \nabla v_i + \nabla v_i^T\right). \label{eq:strainApprox}
\end{equation}
\end{linenomath*}
While the entire strain tensor is computed, we will \change{visualize the maximum principal strains in our results}.

\change{To make our approach more robust to small variations stemming from satellite view angles and lighting conditions, we apply an ice mask to our images and then apply a common technique called Contrast Limited Adaptive Histogram Equalization from the OpenCV Python package.  Ice pixels were identified as pixels with an intensity larger than 120.  For other types of imagery, such as SAR, more complicated techniques may be required to identify ice pixels.
}

\section{Results}
To test our approach and illustrate both its strengths and limitations, we have employed a combination of tests with synthetically generated images as well as MODIS imagery of sea ice in the Nares Strait off the northern coast of Greenland.  Section \ref{sec:results:synthetic} presents the synthetic results and Section \ref{sec:results:nares} presents results for Nares Strait.

\subsection{Synthetic Tests}\label{sec:results:synthetic}

\begin{figure}[!htbp]
\centering

\iftoggle{usetikz}{%
    \include{Figures/SyntheticResults} 
}{%
    \includegraphics{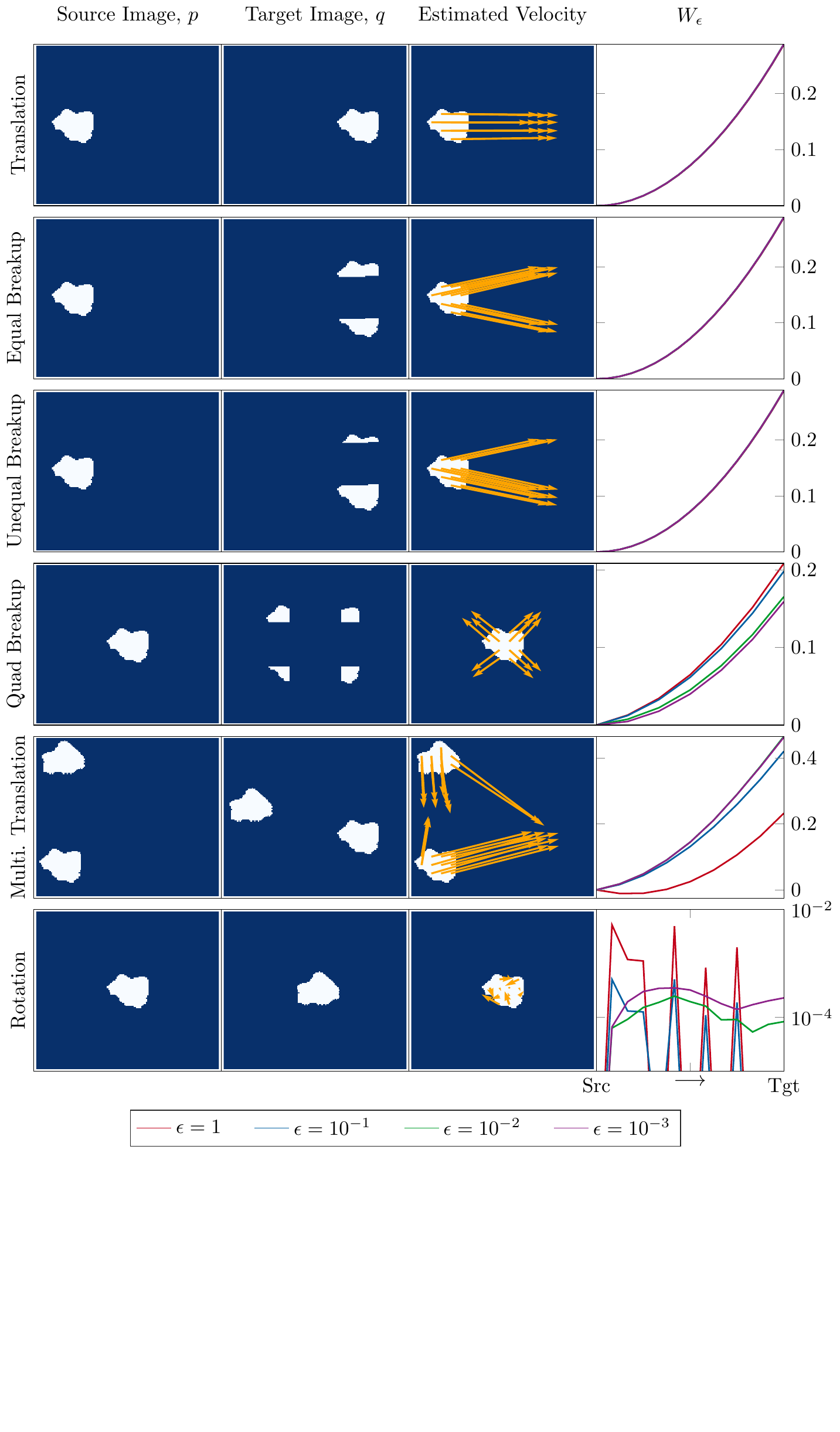}
}
      \vspace{-5.5cm}
  \caption{Source image, target image, and the velocities estimated from \eqref{eq:velocityApprox}.  \textbf{Cases:} (Row $1$) a single floe translating horizontally, (Row $2$) a single floe splitting in half and translating, (Row $3$) a single floe splitting in 20\% and 80\% segments and translating, (Row $4$) a single floe splitting into four segments while maintaining the original average floe position, (Row $5$) two floes translating different amounts in different directions, and (Row $6$) a single floe rotating in-place.}
  \label{fig:MultiFloeTranslation}
\end{figure}

A single ice floe can evolve in many ways: it can translate, rotate, or break up. \change{To understand how the Wasserstein metric responds to each of these scenarios, we have employed several different transformation of a synthetic binary floe.} In Figure \ref{fig:MultiFloeTranslation}, the rows show the different transformation cases, and the columns show, going from left to right, the source image, the target image, velocity, and the trends of $W_\epsilon$ \change{as the floe is transformed. Note that the backgrounds of these test images did not have zero intensity, a small constant value ($\approx 10^{-10}$) was added to the \change{images to ensure they} are nonzero everywhere. The velocity results were calculated using equation \eqref{eq:velocityApprox}, and the relaxation parameter $\epsilon$ was $10^{-3}$ for all velocity images.} Note that the transformation defined in \eqref{eq:transportPlan1}-\eqref{eq:transportPlan2} is defined everywhere, but \change{for clarity we have restricted our attention to the floe}. \change{We parameterized the transformations in each case by a synthetic time variable, $t$, such that $t=0$ results in the source image and $t=1$ results in the target image. The plots in the right column show $W_\epsilon$ as a function of $t$, with the first $W_\epsilon$ value subtracted from each curve.} 

In the floe translation and breakup cases, we found that the Wasserstein distance \change{monotonically increases regardless of the value of $\epsilon$, except for the multiple floe translation case where $\epsilon=1$. In this case, the $W_\epsilon$ curve follows a similar shape as the other cases, but is initially negative. In the simple translation, equal breakup, and the unequal breakup cases, the regularization parameter $\epsilon$ seem to have a negligible impact on the results. However, in the quad breakup, multiple floe, and rotation cases, the Wasserstein distance is over- or under-estimated for the larger values of $\epsilon$. These fluctuations in the rotation case are likely a function of the floe shape, where some features of the rotated floe resemble the initial floe. However, as $\epsilon$ decreases the results for all of these cases converge to a common value, as expected. This illustrates the importance of smaller $\epsilon$ values for finding an optimal coupling between the source and target images. Note that a relaxation parameter of  $\epsilon=1$ is quite large and would not typically be used in practice. However, we also note that larger values of $\epsilon$ result in faster convergence of the iteration in \eqref{eq:sinkhornIterates} and it is necessary to balance accuracy against computation speed in the selection of  $\epsilon$ value.   In both the rotation and multiple floe translation cases, the estimated velocities do not always reflect the true ice motion.  This becomes less of an issue as the deformation between source and target image decreases, which indicates that, as we would expect, obtaining images at small time intervals (compared to ice velocities) is important in the accuracy of the results.  At any rate, these synthetic results indicate that the Wasserstein distance itself $W_\epsilon$ is still a valuable metric for image comparison even under  rotation or breakup scenarios.
}

\subsection{Nares Strait MODIS Imagery}\label{sec:results:nares}
\begin{figure}[!htbp]
	\centering
\iftoggle{usetikz}{%
    \input{Figures/NaresChange}
}{%
    \includegraphics{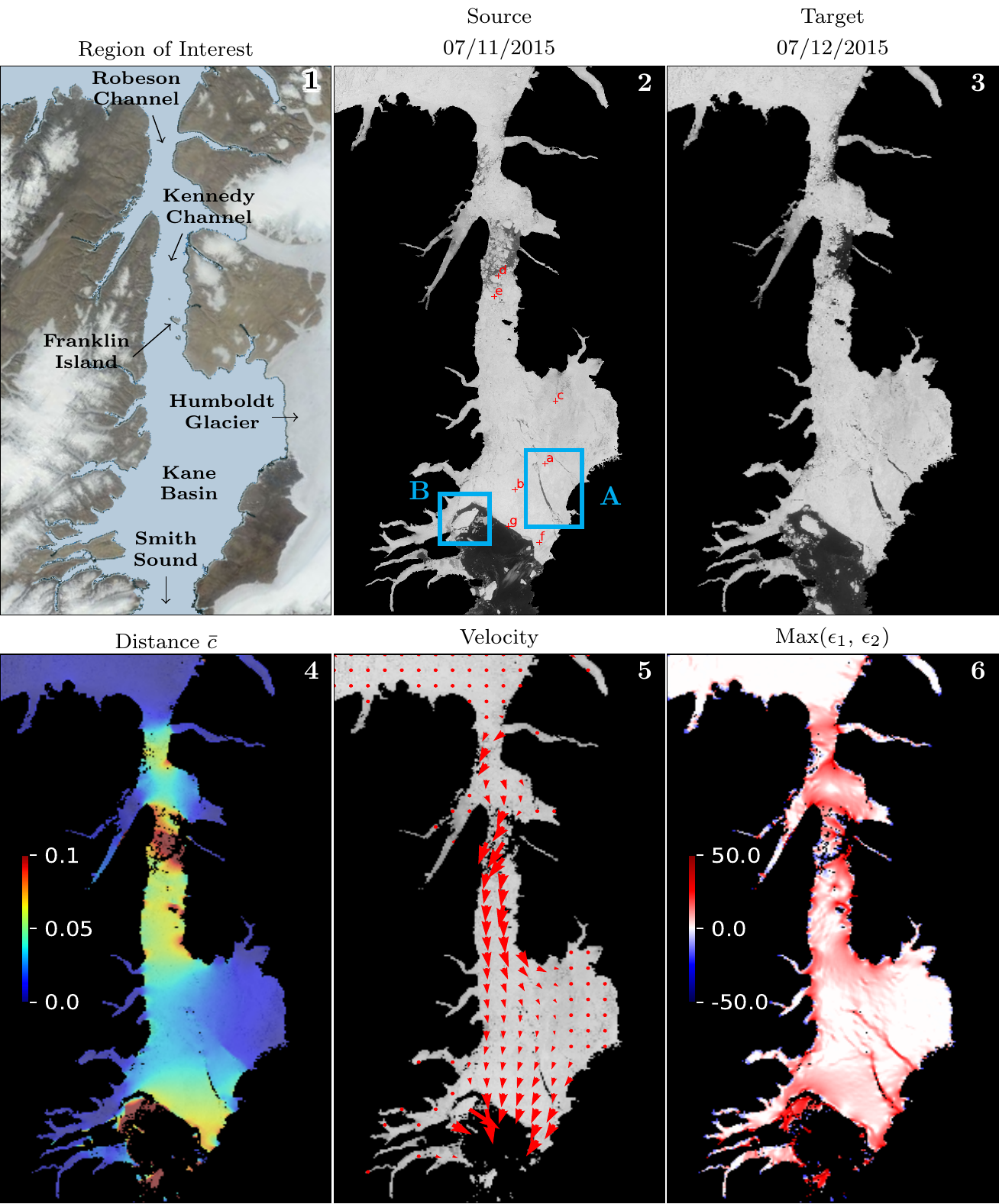}
}
\caption{\change{MODIS imagery for July 11th and 12th, 2015 in the Nares Strait used to demonstrate effectiveness of our approach on real images.  \change{Areas of interest include Robeson Channel, Kennedy Channel, Franklin Island, Humboldt Glacier, Kane Basin, and Smith Sound.  The transport distance $\bar{c}$ has units of m/s, and the compressive (red) and tensile (blue) maximum principal strains were truncated at $\pm50$. Pixels with values less than 120 were masked out for all three results to remove water pixels.  Velocity vectors were thinned by 60 to ease visualization.  Lowercase letters in the source image denote the locations where features were manually tracked for comparison.}}}
\label{fig:NaresStraightImagery}
\end{figure}

One of our goals is to automatically quantify ice deformation as the ice undergoes complicated, spatially variable, combinations of rotation, translation, and breakup.  Sea ice in the Nares Strait, off the northwest coast of Greenland, provides an interesting region to test our approach; as shown in Figure \ref{fig:NaresStraightImagery}, ice in the Nares Strait exhibits complex behavior, where different regions can simultaneously experience fracture, translation, and rotation. In addition, the Strait's geometry, with a wide expanse feeding into a narrow constriction, induces regularly reoccurring fracture patterns, such as the arching fracture patterns that form near the entrance of the Strait \change{\citep{kwok2010large, hibler2006sea}}. Therefore these recognizable reoccurring features provide a qualitative check to verify the utility of the Wasserstein distance metric. 

For our test, we used MODIS-Aqua corrected reflectance imagery for a two day cloud-free period starting on July 11, 2015.  The data were reprojected into polar sterographic coordinates, and land was masked out using shoreline information from the Global Self-consistent, Heirarchical, High-resolution Shoreline Database (GSHHS), \change{leaving $N=709,455$ ocean and ice pixels}. These pixels make up the source and target distributions, $p$ and $q$, in Figure \ref{fig:NaresStraightImagery}. The remainder of Figure \ref{fig:NaresStraightImagery} shows the transport distance, \change{velocity, and maximum principal strain computed} using our optimal transport approach.

\begin{figure}[!htbp]
	\centering
\iftoggle{usetikz}{%
    \input{Figures/VelocityComparison}
}{%
    \includegraphics{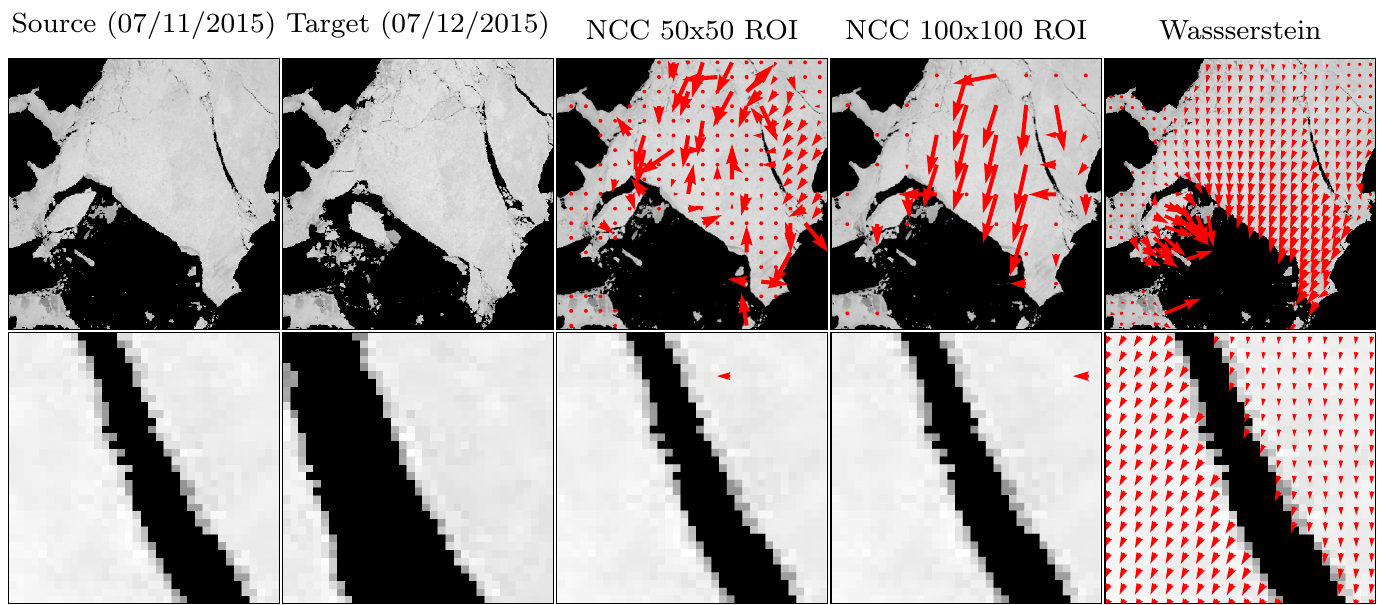}
}
\caption{\change{Comparison of velocity results from the NCC (0.25 correlation threshold) and Wasserstein methods for regions of interest in Kane Basin. The bottom row focuses on a region of the large crack labeled A in the source image of Figure \ref{fig:NaresStraightImagery}. Due to the downsampling effect of the NCC method and the lack of distinct features with the ice, neither result provides much information in the vicinity of this growing lead. However, the Wasserstein method is able to capture the widening crack at pixel resolution (the vector arrows in the top row were thinned by 20 and the bottom row by 2.}}
\label{fig:VelocityComparison} 
\end{figure}

\change{The transport distance $\bar{c}$ and velocity results show a general downward flow of ice through the channels and Kane Basin, which is typical for the Nares Strait. The largest transport costs are observed in Robeson and Kennedy Channels, Kane Basin, and for several individual floes in free drift in the lower part of Kane Basin. In the Kennedy Channel, the highest $\bar{c}$ region corresponds to pieces of ice that consolidated against the western edge of the channel as the ice moved across the channel and downward. In Kane Basin, the transport distance highlights two visibly growing leads (labeled A in the source image) as sharp transitions in $\bar{c}$. These abrupt changes in $\bar{c}$ indicate that each side of the cracks moved independently from each other between July 11 and 12. Our optimal transport method also captured large transport distances for the individual floe (labeled B in the source image), which rotated, translated, and partially broke up. 
The velocity vectors were thinned for visualization purposes, but results for refined regions including this large floe are shown in Figure \ref{fig:VelocityComparison}. These $\bar{c}$ and velocity observations suggest the Wasserstein distance is sensitive to dynamic activity that is difficult to visually discern in the July 11 and 12 image pair alone; a powerful demonstration of this metric's utility.}

\change{The maximum principal strains computed from the transport distance are perhaps even more interesting than the distance and velocity results. 
The two cracks within Kane Basin are clearly visible in the strain plots, which indicates a significant amount of tensile strain as those leads widened. There are also interesting strain patterns in the Robeson and Kennedy Channels and along the islands, notably the concentration of strain North of Franklin Island and the region of low strain in its wake. A strain pattern resembling an arch is also visible near the top of Kennedy channel.   As we would expect, many of these interesting strain features are related to constrictions in the channel, or sub-channels, or where pieces of land extend into the ice. This sensitivity, combined with the interpretability of the strain fields, demonstrates that our technique has great potential for measuring sea ice dynamics.}

\change{For comparison, we also estimated deformation with the NCC technique. Figure \ref{fig:VelocityComparison} compares the NCC results with our approach. To illustrate the dependence of the NCC method on window size, we computed the results for two ROI's (50x50 and 100x100 pixels).} 
\change{The 50x50 case captured small motion such as the piece of ice between the two leads, but failed to capture the large sheet of ice in the middle of the domain moving downward, which was captured in the 100x100 case. These are direct results of the windowing process used in the NCC method, which is problematic for situations where the size of features is unknown, or when estimates are required at a range of length scales. Due to the coarse resolution of the NCC results, neither case provides useful information around the widening lead or the the large individual floe, which were important features in the ice development between July 11 and 12. However, the Wasserstein method was able to capture the transformation of both features at the native scale of the original images. To test these methods further, we compared them against hand-calculated distances of several obvious features designated by lower case letters in the source image of Figure \ref{fig:NaresStraightImagery}.   The features consisted of persistent high pixel intensity floes that could be identified in both images.  We identified pixels in the source and target images containing the same part of these floes and then computed the distance between the source pixel and target pixel.   Repeating this manual identification process several times for the same feature allowed us to estimate a standard deviation of about $500$m in our manual estimates.  The median absolute error between these manual results and NCC was $2304$m and for our optimal transport approach it was $869$m.  Our optimal transport results are within two standard deviations of the manual results and therefore seem to produce reasonable estimates of the actual ice velocity at the test locations.

The pixel-scale resolution of the Wasserstein approach means it is able to calculate accurate displacements for features as small as a few pixels, or features significantly larger, as shown by the synthetic floe and Nares Strait cases. However, we note that large domains with many pixels will increase the time and computational cost of computing the Wasserstein results.   Our approach can in theory be applied to images with any temporal separation.  However, as demonstrated in the synthetic results, in order to produce meaningful results it is important to choose intervals that are short enough to prevent significant amounts of melting, freezing, or mass advecting between the images. This limitation is discussed further in the Conclusions section.}


\section{Conclusions}\label{sec:conclusion}
We have introduced a new method using regularized optimal transport to measure dynamic properties of sea ice like velocity and strain. Our approach treats images as probability distributions and constructs an optimal coupling between the images. This coupling defines a transformation that, unlike existing approaches that are based on cross correlation or image tracking, can naturally handle topology changes that occur during fracture and breakup. Moreover, our approach does not use any form of \change{implicit} homogenization \change{or smoothness assumptions} and can therefore deliver information at the same resolution as the original imagery; thus providing a powerful new way of analyzing remote sensing imagery of sea ice.

\change{
One of the main limitations of our current formulation is that it assumes the sum of the pixel intensities is constant between the source and target images. However, the amount of ice can change due to freezing or thawing, or by advecting into or out of the image domain, which will break this assumption.   In practice, this seems to have the largest impact on velocity estimates near the boundaries, but could theoretically impact estimates over the entire image.  The basic solution to this, which used in our Nares Strait example, is to choose a domain that minimizes the flux of ice across the boundaries.  Such boundaries may have small ice velocities (such as the top of our Nares Strait domain), be over land (the sides of our Nares Strait example), or over relatively ice-free water (the bottom of our example).  }\morechange{These types of boundaries occur in restricted areas like the Nares Strait but can also be found near land fast ice and in Lagrangian image sequences.}   \change{A more advanced solution to this mass conservation assumption could involve quantifying the mass advected through the boundaries, which would result in an unbalanced optimal transport problem.  Unbalanced problems are area of ongoing research in the optimal transport community \citep{chizat2015unbalanced,chizat2018scaling} and any advancements made in their solution could easily be employed within our framework for estimating ice deformation.

Our comparisons with NCC and manual feature tracking provide an initial verification that our approach can accurately characterize ice deformation, but a more detailed comparison with in situ observations is needed will help validate the approach.  Drifting buoys provide a possible data source for this type of validation.    The application of our framework to SAR imagery would also enable a more comprehensive comparison with existing velocity products.  Some additional work will be needed to identify ice pixels in the SAR context, but we do not foresee any fundamental challenges preventing the use of our optimal transport framework with SAR data.

While future efforts will undoubtedly improve our approach, we believe the use of optimal transport, as introduced in this work, has the potential to provide critical high resolution information about dynamic relationships in sea ice.  Indeed, the use of our approach on large datasets of high resolution imagery could provide valuable insight into sea ice dynamics and aid the validation and development of high fidelity sea ice models.

}

\acknowledgments
The work described in this document was funded under the US Army Basic Research Program under PE 61102, Project T22, Task 02 ``Material Modeling for Force Protection" and was managed and executed at the US Army ERDC.

All imagery used in this work is freely available from NASA through their Worldview portal (\texttt{https://worldview.earthdata.nasa.gov}).






%
%
%
%
\bibliography{WassersteinSeaIce}
%
%
%
%
%
%





\listofchanges

\end{document}